\pdfoutput=1

\documentclass[11pt]{article}

\usepackage[final]{acl}

\usepackage{times}
\usepackage{latexsym}

\usepackage[T1]{fontenc}

\usepackage[utf8]{inputenc}

\usepackage{microtype}

\usepackage{inconsolata}

\usepackage{graphicx}
\usepackage{booktabs}
\usepackage{inconsolata}

\usepackage{latexsym}
\usepackage{amssymb}
\usepackage{amsmath}
\usepackage{multirow}
\usepackage{graphicx}
\usepackage{enumitem}

\title{ARMADA: Attribute-Based Multimodal Data Augmentation}

\author{Xiaomeng Jin$^{\;\;\dagger}$ \;  Jeonghwan Kim$^{\dagger}$ \;    Yu Zhou$^{\mathparagraph}$ Kuan-Hao Huang$^{\mathsection}$ \\
\textbf{Te-Lin Wu$^{\ddagger}$ \; Nanyun Peng$^{\ddagger}$ \; Heng Ji$^{\dagger}$} \\[0.5em]
  $^{\dagger}$University of Illinois, Urbana Champaign\; $^{\mathparagraph, \ddagger}$University of California, Los Angeles\; \\ $^{\mathsection}$Texas A\&M University \\
  \texttt{\{xjin17, jk100, hengji\}@illinois.edu}, \quad \texttt{yu.zhou@ucla.edu} \\ 
  \texttt{\{telinwu, violetpeng\}@cs.ucla.edu}, \quad \texttt{khhuang@tamu.edu}}

\begin{document}
\maketitle
\begin{abstract}
In Multimodal Language Models (MLMs), the cost of manually annotating high-quality image-text pair data for fine-tuning and alignment is extremely high.
While existing multimodal data augmentation frameworks propose ways to augment image-text pairs, they either suffer from \textit{semantic inconsistency} between texts and images, or generate unrealistic images, causing \textit{knowledge gap} with real world examples.
To address these issues, we propose \textbf{A}tt\textbf{r}ibute-based \textbf{M}ultimod\textbf{a}l \textbf{D}ata \textbf{A}ugmentation (ARMADA), a novel multimodal data augmentation method via knowledge-guided manipulation of visual attributes of the mentioned entities. Specifically, we extract entities and their visual attributes from the original text data, then search for alternative values for the visual attributes under the guidance of knowledge bases (KBs) and large language models (LLMs). We then utilize an image-editing model to edit the images with the extracted attributes.
ARMADA is a novel multimodal data generation framework that: (i) extracts knowledge-grounded attributes from symbolic KBs for semantically consistent yet \textit{distinctive} image-text pair generation, (ii) generates visually similar images of disparate categories using neighboring entities in the KB hierarchy, and (iii) uses the commonsense knowledge of LLMs to modulate auxiliary visual attributes such as backgrounds for more robust representation of original entities.
Our empirical results over four downstream tasks demonstrate the efficacy of our framework to produce high-quality data and enhance the model performance. This also highlights the need to leverage external knowledge proxies for enhanced interpretability and real-world grounding.

\end{abstract}

\section{Introduction}

Multimodal Language Models (MLMs) exhibit remarkable abilities in comprehending and integrating various modalities, encompassing texts, images, and videos.
Recently, many MLMs have been proposed by researchers in both academic and industrial communities \citep{LiACL2020,radford2021learning, li2022blip, clipevent2022, liu2023visual, dai2023instructblip, achiam2023gpt}, demonstrating significant achievements across various downstream tasks, such as image-text retrieval \citep{radford2021learning, li2022blip} and visual question answering (VQA) \citep{liu2023visual, liu2023improved, dai2023instructblip}. Training MLMs for downstream tasks, which usually involves fine-tuning and alignment stages, requires substantial amounts of annotated data.
However, collecting and annotating such datasets demand considerable human effort and are notorious for their expense and time-consuming nature. A common strategy to overcome this problem is leveraging data augmentation techniques, which automatically synthesize new data instances from existing datasets, relieving the need to rely on manually annotated datasets to train these models.

\begin{figure*}[tb]
  \centering
  \includegraphics[width=0.9\textwidth]{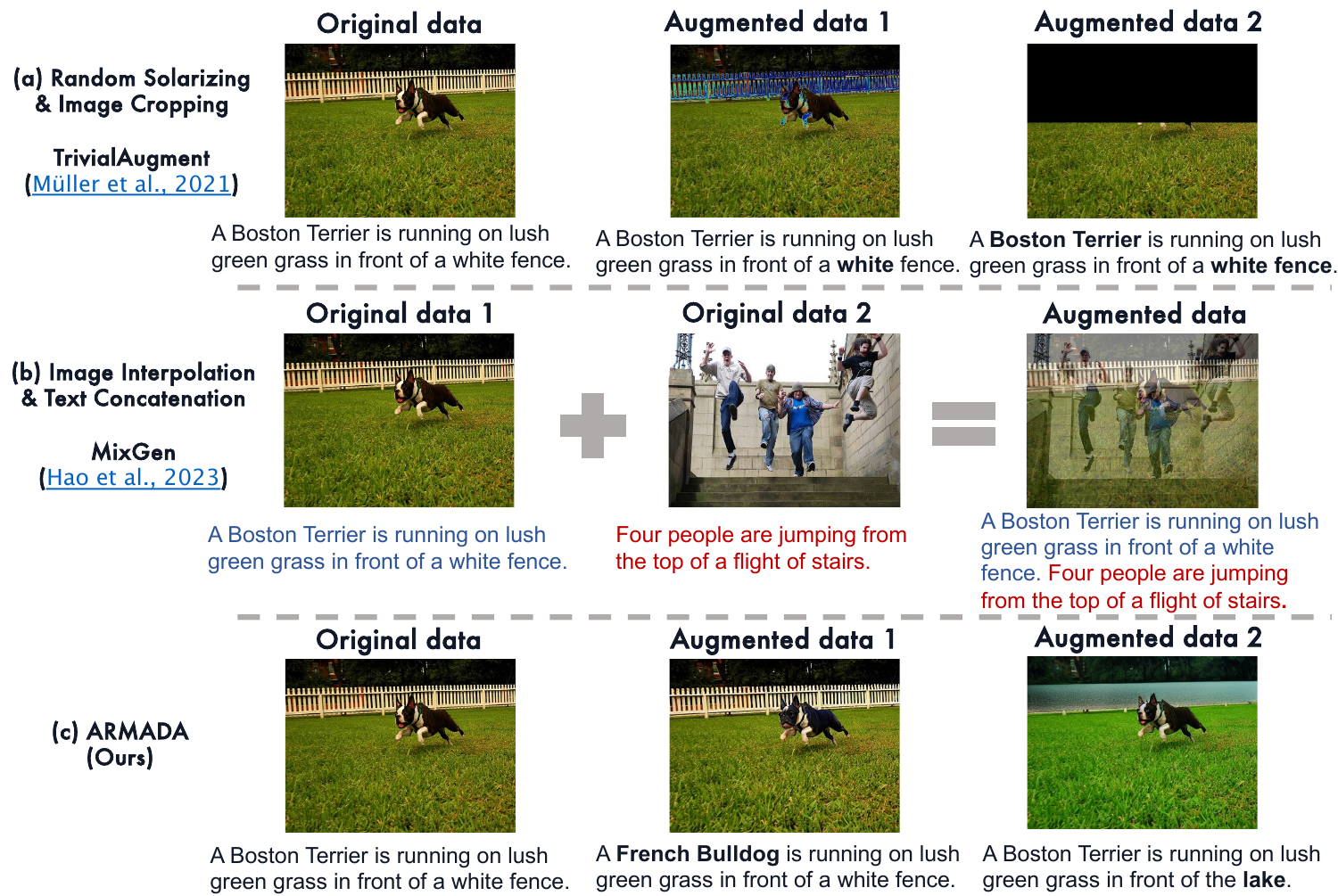}
  \caption{Generated examples using two previous data augmentation methods and our approach. \textbf{(a)} is generated by TrivialAugment \citep{muller2021trivialaugment}, showing the altered images from randomly solarizing or cropping the dog and the fence out from the original image, demonstrating semantic inconsistency. \textbf{(b)} shows the output image from MixGen \citep{hao2023mixgen}, demonstrating the unrealistic output from simple image interpolation and text concatenation.
  \textbf{(c)} shows the augmented data from our method \textit{ARMADA}, which are semantically consistent.
  }
  \label{fig:intro_examples}
  \vspace{-0.5em}
\end{figure*}

Existing multimodal data augmentation methods, which require the perturbation of both the visual and textual modalities in tandem, can generally be classified into the following two groups: (i) latent space-based methods that perturb the latent representations of existing data instances \citep{liu2022learning} via adversarially trained augmentation networks, and (ii) surface form-based methods \citep{muller2021trivialaugment,hao2023mixgen} that simply perturb superficial representations such as orientations/pixel-level mixture of images. Latent space-based methods such as LeMDA \citep{liu2022learning} generate augmented multimodal latent features aligned with the training data distribution, but are inherently confined by their lack of interpretability and controllability. While surface form-based methods partly provide interpretable and controllable alternative, their simple augmentation schemes such as random solarization and pixel-level interpolation lead to \textit{semantic inconsistency}. For instance, Figure \ref{fig:intro_examples} shows that random cropping or image interpolation cause semantic gaps between paired images and texts, leading to images far from realistic. Moreover, such perturbations cannot deal with variable entity categories that appear in a similar background, or same entities with variable physical attributes, since they disregard attribute-level details. 
Our work aims to address these issues by leveraging a rich bank of attributes from a hierarchical knowledge base for interpretable and controllable multimodal data augmentation that guarantees semantic consistency and knowledge-grounding of generated entities.

In this paper, we introduce a novel \textit{attribute-based}, multimodal data augmentation framework, \textit{ARMADA}, that extracts the entities and visual attributes, then modifies the visual attributes of entities in images by building an entity-attribute multimodal knowledge base (KB). We perform entity-related knowledge extraction through entity linking using Spacy Entity Linker on Wikidata KB to: (i) generate augmented images and texts that faithfully reflect knowledge-grounded, entity-related attributes, and (ii) exploit the neighboring entities, e.g., a Boston Terrier and French Bulldog in Figure \ref{fig:intro_examples}, for generating similar yet distinguished entity categories. Our work also leverages LLMs as additional knowledge proxy as they can generate alternatives to any textual attributes without related entities in KB.
We then modify images based on revised texts by employing an off-the-shelf image editing model, InstructPix2Pix \citep{brooks2023instructpix2pix}. Our framework produces semantically consistent, knowledge-grounded multimodal data instances. In-depth experiments across four different image-text downstream tasks against five different baselines demonstrate the significance of augmenting multimodal data instances guided by entity-related attribute knowledge.
Our contributions can be summarized as follows:
\begin{itemize}
    \item We propose a knowledge-guided multimodal data augmentation framework that is guided by entity-centric KBs to generate entities that are of the same type yet differing attributes, or of similar yet disparate categories.
    \item The proposed augmentation pipeline in this work demonstrates semantically consistent and knowledge-grounded multimodal data, addressing the limitations of previous multimodal data augmentation methods.
    \item Our empirical results demonstrate that our proposed data augmentation strategy leads to substantial gains in various image-text downstream tasks such as image-text retrieval, VQA, image captioning, and especially in fine-grained image classification tasks that rely on attribute-centric information.
\end{itemize}

\label{sec:intro}

\section{Related Work}
\textbf{External Knowledge Proxies.}
External symbolic knowledge bases (KBs) like Wikidata \citep{wikidata2014} and real-world knowledge proxies like large language models (LLMs) \citep{achiam2023gpt, touvron2023llama, falcon40b} contain ample amount of real-world, entity-centric knowledge.
While symbolic KBs have frequently been used in various domains of natural language processing for augmentation \citep{luo2023reasoning, sun2023think, pan2024unifying}, the use of symbolic KBs in the multimodal domain is yet to be explored. LLMs, while they may suffer from hallucinatory outputs, contain rich world knowledge that enables them to generalize to attributes of various kinds. Our work reaps the benefits of the both worlds by exploiting the relational knowledge of KBs and generalization abilities of LLMs to perform knowledge-guided multimodal data augmentation.

\noindent \textbf{Vision Language Models.}
Vision Language Models (VLMs) have achieved new state-of-the-art performances across various downstream tasks such as image-to-text retrieval and visual question answering (VQA) \citep{radford2021learning, li2022blip, dai2023instructblip, liu2023visual, liu2023improved}. CLIP \citep{radford2021learning} is a widely used VLM for image-text retrieval and image classification. 
InstructBLIP \citep{dai2023instructblip} and LLaVA \citep{liu2023visual} are instruction-tuned multimodal models that combine vision encoders and LLMs.
The major drawback of these models is that they require an extensive amount of image-text pair datasets to either pre-train or fine-tune the models. Such shortcomings call for the need of a new, robust augmentation method, which our work aims to offer.

\noindent \textbf{Data Augmentation.} 
Existing work on data augmentation mainly focuses on augmenting a single modality, e.g., text \citep{thakur2021augmented, yoo2021gpt3mix, chen2023minprompt} or image \citep{luo2023camdiff, trabucco2023effective, müller2021trivialaugment}.
Most recently in the multimodal domain, several augmentation methods have been proposed to augment multiple modalities at the same time. MixGen \citep{hao2023mixgen} generates new data instances by interpolating images and concatenating their accompanying texts. As discussed in Figure \ref{fig:intro_examples}, one potential issue is the low quality of the generated data.
LeMDA \citep{liu2022learning}, another augmentation method that jointly augments multimodal data in the feature space, is limited in terms of interpretability and controllability since the generation occurs in latent space.
BiAug \citep{wu2023reporting} augments multimodal data in a similar manner as our approach by decoupling entities and their attributes. However, BiAug heavily relies on LLMs to generate the attributes, which are susceptible to hallucinatory outputs. Our proposed approach, in contrast, leverages entity-related attributes from knowledge base and delegates entity independent perturbations to LLMs.


\section{Our Approach}

\begin{figure*}[tb]
  \centering
  \vspace{-0.5em}
  \includegraphics[width=0.9\textwidth]{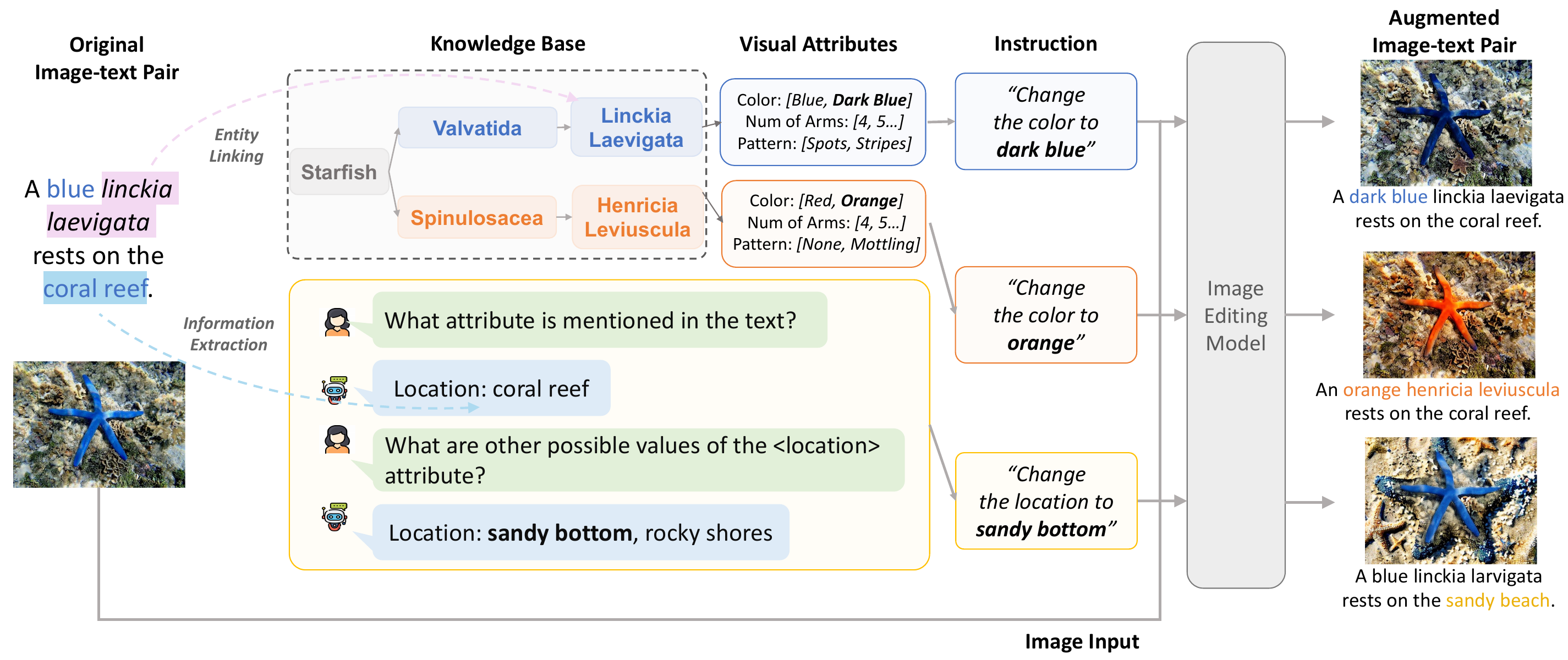}
  \caption{The overall framework of our data augmentation method. Given an image-text pair as input, we first extract entities and their corresponding visual attributes from text. If the object can be linked to an entity in our pre-defined attribute knowledge base, then we collect all possible attribute values from the information of the linked entity. If the object cannot be linked to the knowledge base, then we utilize Large Language Models (LLMs) to extract other possible values. After selecting which visual attribute to modify, we rewrite the original text and use an image editing model to generate new images based on the new text. Finally, we rank the augmented data and output data based on the similarity scores.
  }
  \label{fig:architecture}
\end{figure*}

Suppose we have a set of image-text pairs $\mathcal D = \{(I_1, T_1), \cdots \}$ as the training dataset.
$T_i$ is a task-dependent text that is paired with its corresponding image, $I_{i}$.
For example, $T_i$ can be the label of image $I_i$ in image classification task, a caption that describes $I_i$ in image-text retrieval task, or a question-answer pair if the image $I_i$ appears in a VQA task.
Given that the training dataset with gold-standard annotations $\mathcal D$ is usually too small to train the vision language model sufficiently well, we aim to augment the original training dataset and generate additional image-text pairs $\mathcal D' = \{ (I'_1, T'_1), \cdots\}$.
The augmented dataset $\mathcal D'$ can be used in conjunction with the original dataset $\mathcal D$ to train the VLMs and further improve their performance.


\subsection{Extracting Entities and Visual Attributes from Text}
The primary goal of our proposed data augmentation framework is to generate new images by modifying the value of visual attributes of the mentioned entities.
For example, as shown in Figure \ref{fig:architecture}, our data augmentation method changes the \textit{color} (visual attribute) of a \textit{linckia laevigata} (entity) from \textit{blue} (attribute value) to \textit{orange} (attribute value). The first step of text modification is to identify the mentioned entities and visual attributes of mentioned entities within a given piece of text.
To this end, we use large language models (LLMs) to extract entities, visual attributes and attribute values given an input text, $T$, as they demonstrate exceptional capabilities in text comprehension and generation. Given an original image-text pair $(I, T)$, we input the text $T$ into an LLM along with the prompt ``\texttt{Extract the mentioned objects, their visual attributes, and values of visual attributes from the sentence: $T$}''.
For example, as illustrated in Figure \ref{fig:architecture}, we can extract from the sentence ``\textit{A blue linckia laevigata rests on the coral reef}'' that the \textit{entity} is \textit{linckia laevigata}, the \textit{visual attributes} are \textit{color} and \textit{location}, and the \textit{attribute values} are \textit{blue} and \textit{coral reef}, respectively.
The entities, visual attributes and their values serve as candidates for subsequent visual attribute value substitution.

\subsection{KB-based Visual Attribute Substitution}
\begin{figure*}[tb]
  \centering
  \vspace{-0.5em}
  \includegraphics[width=0.9\textwidth]{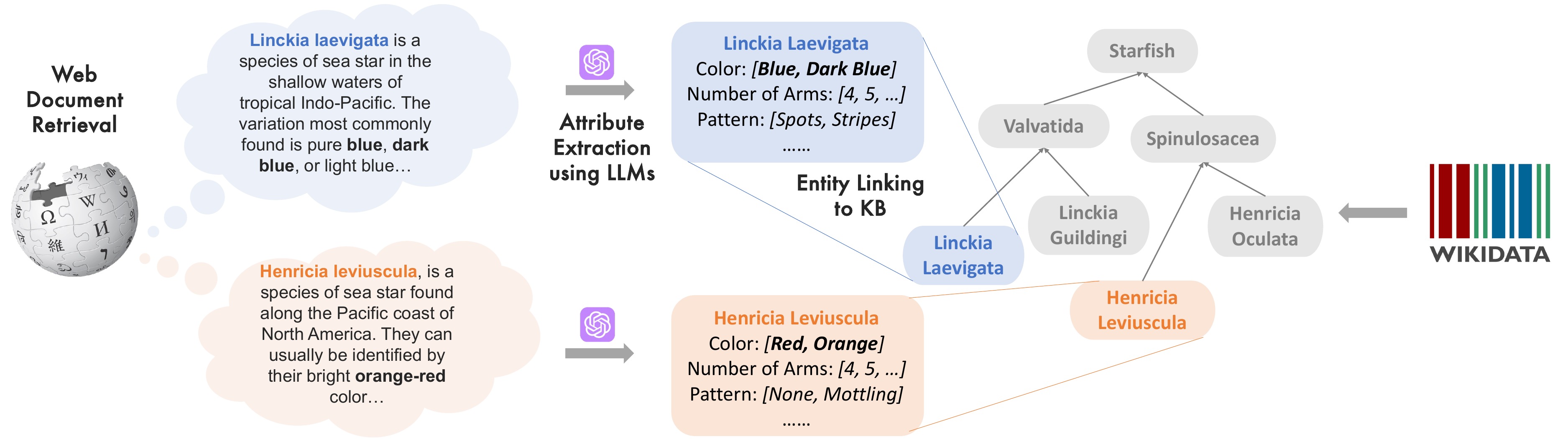}
  \caption{An example from the our pre-defined attribute library. Each node represents an entity collected from Wikidata. An outgoing edge is connected from a node to its parent category. Each node has its visual attributes extracted from the Wikipedia articles.
  }
  \label{fig:attr_lib}
\end{figure*}

\textbf{Knowledge Base Construction.} After identifying visual attributes mentioned in text $T$ we determine potential substitutions for their attribute values. We leverage attributes from entity-centric KBs to provide accurate and reliable knowledge for substituting visual attribute values.
We first parse the information from Wikidata and Wikipedia, and construct an attribute-level KB consisting of entities and their attributes, which consists of two steps:
(1) \textit{Graph topology}: We collect entities from Wikidata and use a node in the KB to represent an entity.
Each node has an outgoing edge to its parent category node.
For instance, as illustrated in Figure \ref{fig:attr_lib}, both \textit{linckia laevigata} and \textit{linckia guildingi} belong to the parent category \textit{valvatida}, thus resulting in two directed edges from these nodes to \textit{valvatida}.
(2) \textit{Node attributes}: The visual information for each node in the KB is derived from its corresponding Wikipedia articles. We collect the textual content of each Wikipedia page, then employ LLMs to extract all visual attributes and their possible values described within the article.
For instance, the entity \textit{linckia laevigata} may have \textit{color} of \textit{blue} and \textit{dark blue}, with the \textit{number of arms} starting from four.

After building the KB, we link each entity extracted from $T$ to a node $N$ in the KB using the Spacy Entity Linker \citep{honnibal2020spacy}.
To generate a new augmented data sample, we use the following two attribute value substitution methods.

\noindent \textbf{Attribute Substitution within Single Entity.} A single entity may possess multiple plausible attributes, which are identifiable through entity linking to KB. Some of these extracted entities with specific attributes may occur less frequently in the original training dataset than those with more frequently occurring attributes. Therefore, we aim to augment the data to increase the coverage of such long-tail entity instances, so that the model is better fine-tuned to recognize these rare cases well. To elaborate, we randomly choose a visual attribute connected to the entity node $N$ and then sample an attribute value to substitute the current attribute value of $N$.
    In this case, the entity stays the same while only its one attribute value is changed. For example, \textit{blue linckia laevigata} $\rightarrow$ \textit{dark blue linckia laevigata} as illustrated in Figure \ref{fig:architecture}.
    
\noindent \textbf{Attribute Substitution across Sibling Entities.} In addition to substituting attributes within a single entity, we notice that there are many entities in KBs that belong to the same parent category and share many visual attributes in common, e.g., the \textit{linckia laevigata} and \textit{henricia leviuscula} in Figure \ref{fig:architecture}.
    This inspires us to substitute attributes across these sibling entities to introduce similar but different concepts as augmented training data.
    In this way, the model will contrastively learn from these confusing entities, thereby increasing its robustnesss to visually similar but different entity concepts.
    Specifically, we consider changing the entity node $n_{i}$ to its sibling entity node $n_{s}$ who share the most visual attributes with $n_{i}$. 
    For example, in Figure \ref{fig:attr_lib}, \textit{linckia laevigata} and \textit{henricia leviuscula} have many attributes in common, so it is feasible to change the original entity to the new entity.
    We therefore substitute the entity \textit{linckia laevigata} with \textit{henricia leviuscula}, and then change its \textit{color} for \textit{henricia leviuscula} (e.g., \textit{orange}).
    The resulting substitution is therefore \textit{blue linckia laevigata} $\rightarrow$ \textit{orange henricia leviuscula}.

\subsection{LLM-based Visual Attribute Substitution}
In some cases, the extracted entity or visual attribute is too general and cannot be linked to any node in the KB (e.g., \textit{coral reef} serving as a background in Figure \ref{fig:architecture}).
Therefore, in addition to KBs, we also use LLMs to obtain new values for auxiliary visual attributes such as background, as they are broadly trained on a large amount of data and thus have acquired commonsense knowledge to provide alternative attribute values for such cases.
For example, in Figure \ref{fig:architecture}, after we extract that the \textit{location} is \textit{coral reef}, we use the prompt ``\texttt{What are other possible values for the <location> attribute in this sentence?}'' to generate new location value substitutions, such as \textit{sandy bottom} and \textit{rocky shores}.
It is worth noting that LLMs may not consistently produce valid substitute attribute values, as they may lack adequate knowledge regarding specialized fields or long-tail concepts. This deficiency may lead to LLMs generating inaccurate responses, i.e., hallucination. For instance, when prompt the LLMs for all possible colors of \textit{linckia laevigata}, LLMs may provide incorrect answers such as ``\textit{orange}'' and ``\textit{yellow}'', which are implausible colors for \textit{linckia laevigata}. Therefore, we rely on KBs to extract accurate, knowledge-grounded attributes for substitution.

It is worth noting that the models we utilize in each component may not be perfect, which can affect the performance of the proposed approach. Our experimental results in Section \ref{error_analysis} indicate that the error rates of the information extraction, entity linking, and visual attribute substitutions are relatively low, which do not significantly impact the quality of the generated data.

\subsection{Image Editing}
After modifying an image-text pair $(I, T)$ to $(I, T')$ with a new text $T'$, we edit the image $I$ according to $T'$. We employ an image editing model InstructPix2Pix \citep{brooks2023instructpix2pix}, which can take as input an image and instruction on how to modify the image, and output the modified image following the instruction.
The instruction here is ``\texttt{Change the [attribute] of the [entity] to [value]}'', where \texttt{[entity]} and \texttt{[attribute]} are the mentioned entity and selected attribute type, respectively, and \texttt{[value]} is the new attribute value output by the KB or LLM.
As illustrated in Figure \ref{fig:architecture}, starting with the original image on the left, we generate three new images on the right using InstructPix2Pix with different instructions.
The first image keeps the entity \textit{linckia laevigata} unchanged while changing its color to \textit{dark blue}, whereas the second image changes the color to \textit{orange}, updating the entity category to \textit{henricia leviuscula} and its corresponding text description accordingly.
The third one is the result of changing the attribute of \textit{location} to \textit{sandy beach} by querying LLMs; this leaves the central entity of the image unperturbed, providing a robust way to leverage LLMs only for attributes that are not entity-related.

\subsection{Augmented Data Selection}
\label{sec:augmented_data_selection}
Our method transforms an image-text pair $(I, T)$ to a modified image-text pair $(I', T')$.
However, not all modified image-text pairs are suitable as augmented data;
some image $I'$ being too similar to their original counterpart $I$, thereby providing minimal new signal for subsequent model training.
Conversely, other generated image $I'$ diverging too much from their original counterpart $I$ may significantly drift the image away from the original data distribution and mislead the model training.
To determine the validity of the augmented data, we calculate the similarity between a generated image $I'$ and its original image $I$ using the Fréchet Inception Distance (FID) score \citep{heusel2018gans}. FID calculates the Fréchet distance between feature vectors of the original and generated images, which aligns closely with human judgment and is frequently utilized to assess the quality of generated data.
Ideally, we aim to empirically maintain the similarity score within a specific range to ensure that $I'$ exhibit \textit{a reasonable amount} of difference from $I$ as indicated in the ablation study.
The experimental results on selecting the similarity range is presented in Appendix \ref{sec:ablation_sim}.

\section{Experiments}

To assess the effectiveness of data augmentation methods, we select four evaluation tasks: \textit{image classification}, \textit{visual question answering}, \textit{image-text retrieval}, and \textit{image captioning}.

\subsection{Foundation Models and Baseline Methods}
We use CLIP \citep{radford2021learning} and LLaVA-1.5 (7B) \citep{liu2023improved}  model as the foundation models in this work. CLIP is a multimodal model that uses contrastive learning to jointly align the visual and textual representations. 
LLaVA-1.5 is an open-source, auto-regressive multimodal vision-language model (VLM) trained by fine-tuning Vicuna-v1.5 \citep{vicuna2023} on GPT-4-generated multimodal instruction-following data. Given an image input and text instruction, LLaVA-1.5 generates output texts based on its reasoning upon the two modalities.
We use GPT-4 \cite{openai2023gpt} as the LLMs in each component.

We compare our proposed method against five different baseline methods to demonstrate its effectiveness (we do not include BiAug \citep{wu2023reporting} since the code has not been released yet):
(1) \textit{Zero-shot}: Models are evaluated without fine-tuning on any data. This setting is established to examine the initial ability of the models on all four downstream tasks.
(2) \textit{NoAug}: Only the original training data is used to fine-tune the models without any augmented data.
(3) \textit{NaiveAug}: Two naive augmentation methods are applied to texts and images independently as follows.
We use AEDA \citep{karimi2021aeda} to randomly insert punctuation marks into original text, and we use TrivialAugment \citep{müller2021trivialaugment} to randomly apply center cropping, rotation, or invert, to images.
(4) \textit{MixGen} \citep{hao2023mixgen}: Generates new data instance by interpolating images on the pixel-level and concatenating texts. This is state-of-the-art augmentation method. Specifically, given two image-text pairs $(I_i, T_i)$ and $(I_j, T_j)$, a new image-text pair $(I'_k, T'_k)$ is generated by $I'_k = \lambda I_i + (1 - \lambda) I_j$ and $T'_k = concat(T_i, T_j)$, where $\lambda$ is a hyper-parameter.
(5) \textit{LeMDA} \citep{liu2022learning}: Generates augmented data in the latent feature space. We use CLIP to encode the original training data into embeddings, then feed them to LeMDA to generate new latent embeddings; these embeddings are used as augmented data to fine-tune an MLP module in the image classification task. Note that LeMDA cannot be used for LLaVA-1.5 and cannot be used in tasks other than image classification.

\subsection{Image Classification}
\textbf{Dataset.}
We use \textit{iNaturalist 2021} \citep{vanhorn2018inaturalist} as the dataset for image classification.
The iNaturalist dataset consists of large scale species of plants and animals in the natural world.
It contains 10,000 species with a training set of 2.7M images.
To better mimic the scenario of annotated data scarcity, we sample from a mini dataset with all 246 species of Mammalia.
Each class has 30/15/15 images for training/validation/inference.

\noindent \textbf{Experimental Setup.}
For CLIP, we transform the class labels in iNaturalist dataset into natural language descriptions: ``[label]'' $\rightarrow$ ``\textit{a photo of} [label]'', following caption formats in CLIP \citep{radford2021learning}.
CLIP takes as input an image and all class labels, then outputs logit scores for these classes.
The label with the highest logit score is taken as the predicted result of CLIP model.
For LLaVA-1.5, we evaluate its performance by asking the model what is included in the image, and then verify whether the true labels are presented in the generated responses.
The evaluation prompt is: ``\texttt{What is the name of the mammal that appears in this image? For example, if it's a picture of a bengal tiger, output a fine-grained label `Bengal Tiger' or use its binomial nomenclature `Panthera tigris tigris'. Provide your answer:}''.
This allows us to assess model's classification ability based on the provided images.

\begin{table*}[t]
  \centering
  \small
  \setlength{\tabcolsep}{8pt}
  \begin{tabular}{c|ccc|cc||cc}
    \toprule
    \multirow{2}{*}{Method} & \multicolumn{3}{c|}{\textbf{Image Classification (CLIP)}} & \multicolumn{2}{c||}{\textbf{Image Classification (LLaVA)}} & \multicolumn{2}{c}{\textbf{VQA}} \\
    & $Precision$ & $Recall$ & $F_1$ & $F_1$ & $Exact Match$ & $USE$ & $BERTScore$ \\
    \midrule
    Zero-shot  & $0.074$ & $0.113$ & $0.090$ & $0.041$ & $0.002$ & $0.221$ & $0.825$ \\
    NoAug & $0.332$ & $0.347$ & $0.339$ & $0.517$ & $0.557$ & $0.815$ & $0.949$ \\
    NaiveAug & $0.386$ & $0.336$ & $0.359$ & $0.192$ & $0.241$ & $0.821$ & $0.961$ \\
    MixGen & $0.343$ & $0.318$ & $0.330$ & $0.314$ & $0.357$ & $0.824$ & $0.959$ \\
    LeMDA & $0.368$ & $0.354$ & $0.361$ & - & - & - & - \\
    \midrule
    ARMADA & $\textbf{0.391}$ & $\textbf{0.386}$ & $\textbf{0.389}$ & $\textbf{0.588}$ & $\textbf{0.621}$ & $\textbf{0.835}$ & $\textbf{0.975}$ \\
  \bottomrule
  \end{tabular}
  \caption{Results of Precision, Recall, and F$_1$ on iNaturalist dataset for image classification (left part) and results of textual similarity on VQA v2.0 dataset for visual question answering (right part). The foundation model is LLaVA-1.5 for VQA.}
  \label{tab:ic_results}
\end{table*}

\noindent \textbf{Results.} \
The results of \textit{Precision}, \textit{Recall}, and \textit{$F_1$} for image classification task are presented in the left part of Table \ref{tab:ic_results}.
As shown from the zero-shot results, the pretrained foundation models have poor performance on fine-grained concept recognition, with $F_1$ scores of $0.090$ and $0.041$ on CLIP and LLaVA, respectively.
After fine-tuning with the original training data, both models have a much better performance, with a $24.9\%$ and $47.6\%$ absolute gain on $F_1$ scores.
While both NaiveAug and LeMDA demonstrate some improvement in model performance, our method achieves the best results among all existing methods.
It is worth noting that the $F_1$ score of MixGen is worse than NoAug.
This is because the interpolation of images distorts the visual attribute of the fine-grained concepts, thereby adversely affects model training.
Conversely, our method is able to generate new images by modifying the visual attributes of entities.
This facilitates a more comprehensive learning of fine-grained concepts by foundation models.

\subsection{Visual Question Answering}
\textbf{Datasets.}
Visual Question Answering (VQA) v2.0 \citep{balanced_vqa_v2} dataset consists of open-ended questions to images.
These questions require understanding vision, language, and commonsense knowledge to provide answers.
VQA-2.0 has 265,015 images and each image has at least 3 related questions.

\noindent \textbf{Experimental Setup.}
We consider the VQA task as an answer generation task.
We utilize LLaVA as the foundation model.
Given the open-ended nature of the task, we let the model generate free-form answers without any constraints.
Then we compute the textual similarity between the output of LLaVA and the true answer.


\noindent \textbf{Results.}
The results of VQA task are shown in the right part of Table \ref{tab:ic_results}.
We evaluate the performance on the test-dev dataset via textual similarities using Universal Sentence Encoder (USE) \citep{cer2018universal} and BERTScore \citep{zhang2020bertscore}..
It is clear that, compared with Zero-shot, the performance of LLaVA improves greatly after fine-tuning.
This is probably because the ground truth answers to the questions are typically simple and short, which makes the task relatively easier.
As demonstrated in the table, the textual similarity achieved by our method surpasses the best baseline method MixGen by $1.1\%$ on USE and $1.4\%$ on BERTScore.

\subsection{Image-Text Retrieval}
\label{sec:image_text_retrieval}
\textbf{Dataset.}
\textit{Flickr30k} \citep{young-etal-2014-image} contains 31,000 images, each with 5 human-annotated referenced sentences that describe the image.
This dataset is widely used in image-text retrieval task.
Similar to iNaturalist, we sample 5k images from the training set and use the entire 1k test set for evaluation.

\noindent \textbf{Experimental Setup.}
Image-text retrieval includes two subtasks: text-to-image and image-to-text retrieval. We use CLIP to calculate the embedding of the given image, as well as the embeddings of all candidate captions in the test set.
We compare the cosine similarity between the image embedding and each text embedding, and output top $K$ captions with the highest similarity scores as the retrieved results.
We follow existing work and use $Recall@K$ as evaluation metric.

\begin{table*}[tb]
  \centering
  \small
  \setlength{\tabcolsep}{9pt}
  \begin{tabular}{c|ccc|ccc||cc}
    \toprule
    \multirow{2}{*}{Method} & \multicolumn{3}{c|}{\textbf{Image Retrieval}} & \multicolumn{3}{c||}{\textbf{Text Retrieval}} & \multicolumn{2}{c}{\textbf{Image Captioning}} \\
	& $R@1$ & $R@3$ & $R@5$ & $R@1$ & $R@3$ & $R@5$ & \textit{USE} & \textit{BERTScore} \\
    \midrule
    Zero-shot & $0.589$ & $0.765$ & $0.824$ & $0.612$ & $0.775$ & $0.836$ 
& $0.422$ & $0.896$ \\
    NoAug & $0.619$ & $0.785$ & $0.830$ & $0.645$ & $0.807$ & $0.854$ & $0.642$ & $0.907$ \\
    NaiveAug & $0.631$ & $0.788$ & $0.838$ & $0.641$ & $0.804$ & $0.862$ & $0.648$ & $0.911$ \\
    MixGen & $0.626$ & $0.786$ & $0.838$ & $0.592$ & $0.770$ & $0.826$ & $0.659$ & $0.903$ \\
    \midrule
    ARMADA & $\textbf{0.646}$ & $\textbf{0.797}$ & $\textbf{0.847}$ & $\textbf{0.646}$ & $\textbf{0.811}$ & $\textbf{0.872}$ & $\textbf{0.682}$ & $\textbf{0.918}$ \\
  \bottomrule
  \end{tabular}
  \caption{Results of Recall@K for image-text retrieval (left part) and results of textual similarity for image captioning (right part). We use Flickr30k dataset for both tasks. The foundation model is CLIP for image-text retrieval and LLaVA-1.5 for image captioning.}
  \label{tab:itr_results}
\end{table*}

\noindent \textbf{Results.}
The results of image-text retrieval are shown in the left part of Table \ref{tab:itr_results}.
The zero-shot performance of the pretrained CLIP is already very good on both image retrieval and text retrieval, because it is originally trained using the contrastive loss between image and text embeddings.
After fine-tuning, the performance on both subtasks can be further improved in most cases.
Note that the improvement of our method over baseline methods in this task appears less significant compared to other tasks.
This is primarily due to the already high zero-shot performance of CLIP, leaving limited room for further improvement.

\subsection{Image Captioning}
\textbf{Experimental Setup.}
Image captioning task aims to generate natural language descriptions of an image.
We use LLaVA-1.5 as the foundation model, and Flickr30k as the evaluation dataset as introduced in Section \ref{sec:image_text_retrieval}. Specifically, given an image as input, we use the prompt ``\texttt{Describe this image using one simple sentence}'' to ask LLaVA-1.5 to generate a caption.

To evaluate the quality of generated captions, we compare the textual similarity between the generated caption and the gold-standard annotation for a given image using USE and BERTScore.
Since there may be multiple gold-standard captions for an image, we calculate the similarity score of a generated caption with each gold-standard caption, and return the maximum as the final score for this generated caption.

\noindent \textbf{Results.} \
The results of image captioning task are presented in the right part of Table \ref{tab:itr_results}.
Our method \textit{ARMADA} achieves the best performance over all baseline methods.
Specifically, the performance gain of our method on USE score is $4.0\%$ over NoAug, and $2.3\%$ over the best baseline augmentation method MixGen.
We provide detailed case analysis of the generated captions by our method and by baseline methods in Appendix \ref{sec:case_analysis}.

\subsection{Error Analysis}
\label{error_analysis}
We investigate the error rate of each component in the data augmentation process and how they affect our model. Specifically, we manually check the correctness of attribute extraction and the visual attribute substitution.
It turns out that the percentage of incorrect attributes that are extracted is quite low (4 / 113 = 3.5\%). The percentage of inappropriate substitution by LLMs is also very low (1 / 73 = 2.7\%). The visual attribute substitutions from KBs are template-based substitutions from possible attribute values, which will not incur any error aggregation issues.

\section{Conclusions and Future Work}
We propose a novel data augmentation method that utilizes KBs and LLMs to generate multimodal data. The proposed framework is able to generate semantically consistent data that solves the potential issues of the existing methods. Our method significantly improves the MLM’ performance on various downstream tasks, without the need of high-cost annotated data. Experiment results also demonstrate the effectiveness of our proposed method compared to the baseline methods.

In the future, we aim to incorporate more modalities into our framework such as video and audio.
We also plan to rank visual attributes and select the most influential attributes for augmentation.
Moreover, existing image editing tools our framework relies on do not perform consistently well. Designing a new visual attribute editing model to further enhance the quality of the augmented data is also a promising research direction.

\clearpage

\section{Limitations}
Our proposed method demonstrates the effectiveness only on image-text data. However, to enhance the practical utility of our method, it would be advantageous to expand our data augmentation method to include more modalities, such as video and audio. Furthermore, as discussed earlier, although the error rate in each component is low and will not affect the performance much, we still aim to incorporate better attribute extraction and visual attribute substitution models into the framework to further improve our method.

\section{Ethical Consideration}
We acknowledge that our word is aligned with the \textit{ACL Code of the Ethics} \cite{gotterbarn2018acm} and will not raise ethical concerns.
We do not use sensitive datasets/models that may cause any potential issues.

\clearpage

\bibliography{custom}

\clearpage

\appendix
\label{sec:appendix}
\section{Ablation Study}
\label{ablation_study}
\subsection{Impact of the Size of the Generated Data}
To investigate the impact of the amount of the augmented data, we conduct experiments by varying the size of the augmented data relative to the size of the original training data, ranging from 0\% to 300\%.
The results in Table \ref{tab:abl_size} show a decline in model performance when the augmented data size significantly surpassed the original training data size (exceeding $100\%$ to $200\%$), potentially due to excessive noise introduced by the augmented data. Our findings suggest that, the augmented data size should approximate that of the original training data for best performance.

\begin{table*}[h]
  \centering
  \small
  \setlength{\tabcolsep}{13pt}
  \begin{tabular}{c|c|cccc}
    \toprule
    \multirow{2}{*}{Dataset} & \multirow{2}{*}{Metric} & \multicolumn{4}{c}{Size of the augmented data} \\
	& & $0\%$ & $100\%$ & $200\%$ & $300\%$ \\
    \midrule
    iNaturalist & $F_1$ & $0.339$ & $\textbf{0.389}$ & $0.360$ & $0.328$ \\
    Flickr30k & $TextSim$ & $0.642$ & $\textbf{0.682}$ & $0.660$ & $0.659$ \\
    VQA v2.0 & $TextSim$ & $0.815$ & $0.825$ & $\textbf{0.835}$ & $0.816$ \\
  \bottomrule
  \end{tabular}
  \caption{The impact of the size of the generated data on the performance of multiple tasks.}
  \label{tab:abl_size}
\end{table*}

\subsection{Impact of Using KBs}
To assess the importance of utilizing KBs, we conduct additional experiments on the image classification task by solely relying on LLMs to do attribute value substitution.
Following the aforementioned experimental setup, we fine-tune a CLIP model on the iNaturalist dataset.
The $F_1$ score exhibits a $2.5\%$ decline (from $38.9\%$ to $36.4\%$) without using KBs.
This suggests that though LLMs are able to provide answers for attribute value substitution, the hallucination issue on fine-grained or rare entities can still introduce noise to the training data, thereby impacting the model performance.

\subsection{Impact of Similarity Range for Selecting Augmented Data}
\label{sec:ablation_sim}
We conduct experiments to investigate how the similarity between augmented and original data impact the model performance.
In the image classification task, we split the augmented dataset into four groups of equal size according to the similarity of the edited image with its original image.
Then we use each group as the augmented data to train CLIP.
The $F_1$ scores of the four groups are $0.377$, $0.389$, $0.383$, and $0.364$, respectively, from most-similar to most-dissimilar.
The results support our claim in Section \ref{sec:augmented_data_selection} that maintaining similarity scores within a reasonable range achieves the best performance.

\begin{figure*}[b]
  \centering
  \includegraphics[width=\textwidth]{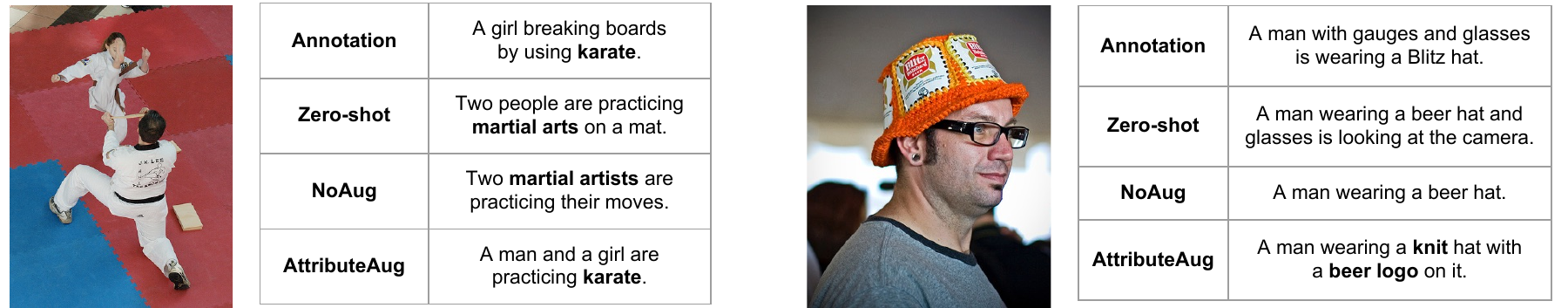}
  \caption{A case analysis that shows sample outputs on Flickr30k dataset for image captioning task. We select two images from the test set, the human-annotated captions, and the generated captions from each method. For the image on the left, our method is able to recognize the fine-grained concept \textit{karate}. The image on the right demonstrates that the model is able to provide an accurate description of the \textit{hat}, specifying its \textit{knit} texture and \textit{beer logo} pattern.}
  \label{fig:case_study}
\end{figure*}

\section{Case Analysis on the Results of Image Captioning}
\label{sec:case_analysis}
We perform a case analysis to illustrate the effectiveness of our method.
In Figure \ref{fig:case_study}, we present two image-caption pairs from the Flickr30k dataset, including both the human-annotated captions and the captions generated by Zero-shot, NoAug, and \textit{ARMADA} (using LLaVA as the foundation model).
For the image on the left, our method is able to identify the fine-grained concept \textit{karate} whereas the zero-shot and NoAug methods generate a more generalized concept \textit{martial arts}.
For the image on the right, the caption generated by our method provides a more detailed and accurate description of the hat, which specifies its \textit{knit} pattern and the \textit{beer logo} pattern.
These examples suggest that LLaVA can effectively learn the visual attributes and identifies the fine-grained concepts through our method.

\section{Ethical Consideration}
We acknowledge that our word is aligned with the \textit{ACL Code of the Ethics} \cite{gotterbarn2018acm} and will not raise ethical concerns.
We do not use sensitive datasets/models that may cause any potential issues.


\end{document}